# Attention Mechanism for Multivariate Time Series Recurrent Model Interpretability Applied to the Ironmaking Industry

Cedric Schockaert[†]
Paul Wurth S.A.
Luxembourg, Luxembourg
cedric.schockaert@paulwurth.com

Reinhard Leperlier
UFR Sciences et Technologie
Université de La Rochelle
La Rochelle, France
reinhardleperlier@gmail.com

Assaad Moawad
DataThings S.A.R.L.,
Luxembourg, Luxembourg
assaad.moawad@datathings.com

## ABSTRACT

Data-driven model interpretability is a requirement to gain the acceptance of process engineers to rely on the prediction of a data-driven model to regulate industrial processes in the ironmaking industry. In the research presented in this paper, we focus on the development of an interpretable multivariate time series forecasting deep learning architecture for the temperature of the hot metal produced by a blast furnace. A Long Short-Term Memory (LSTM) based architecture enhanced with attention mechanism and guided backpropagation is proposed to accommodate the prediction with a local temporal interpretability for each input. Results are showing high potential for this architecture applied to blast furnace data and providing interpretability correctly reflecting the true complex variables relations dictated by the inherent blast furnace process, and with reduced prediction error compared to a recurrent-based deep learning architecture.

## CCS CONCEPTS

• Computing methodologies~Machine learning~Machine learning approaches~Neural networks

## 1 Introduction and background

Deep learning models have recently brought large expectation for different applications in artificial intelligence. Those models have an inherent capability to capture complex relations between large number of variables, structured and unstructured, temporal and/or spatial [1, 2]. The emergence of IoT solutions to monitor an industrial process is generating an enormous quantity of time series data. As an illustration, in the ironmaking industry, the number of sensors for a blast furnace is several thousands, with sensors measuring temperatures, pressures, flows, chemical contents, etc. Analyzing properly those multivariate time series is bringing deep insight about the process itself, triggering understanding for corrective actions to be taken in order to optimize the process, and to reach production key performance indicators (KPIs). For that purpose, deep learning models for hot metal temperature forecast have been developed to guide the blast furnace operator to take appropriate decisions for optimal thermal regulation of the blast furnace [3]. However, to bring the data scientist and the process engineer to a certain level of confidence to deploy and use in production a data driven model predicting the evolution of the hot metal temperature, insight about variables having a high importance for each prediction is required. Indeed, this is often the first major obstacle to reach the successful acceptance by the process expert to rely on data driven model for improved process regulation and leading the journey towards higher level of automation in the ironmaking industry.

Recurrent-based deep learning models are designed to process multivariate time series data having complex temporal and inter time series non-linear relations. For a blast furnace, each input variable has a different reaction time on the hot metal temperature to predict due to the high inertia of the underlying process. Those temporal shifts in the dataset are temporally dynamic as they also depend on the current operation of the blast furnace, which increases the level of complexity to capture causal relations in multivariate time series. Recurrent deep learning models have a dedicated architecture allowing them to model those dynamic causal relations but are acting as black box. Therefore, it is not possible to interpret the predictions of such a model that would require accessing the information about variables and corresponding temporal shifts that have been important to generate each single prediction. This inherent black box characteristic of deep learning models is the price to pay today to develop predictive models with better accuracy compared to conventional machine learning approaches, some being interpretable such as CART [4] or linear Regression. The need to open the box of deep learning models has created a new research field, where the research community is investing considerable efforts to develop interpretable deep learning model [5].

Deep learning model for time series interpretability has been however covered in only few papers. Two approaches are usually proposed in the literature: model-agnostic or model-specific interpretability. In the model-agnostic approach, a dedicated post-analysis of a trained model is performed by generating perturbations in the training data [6], by calculating Shapley values [7], or by analyzing gradient propagation in a network [8]. The model-specific category is aiming at building attention mechanism directly embedded in the Long Short-Term Memory (LSTM) architecture. The attention mechanism is contributing to improve the model accuracy by enhancing using dynamic weight, only relevant information in the recurrent hidden states [9, 10, 11, 12]. Indeed, recurrent networks tend to be less sensitive to pattern occurring far from the current time, which reduces their performances and is known as the vanishing gradient. An improvement to this issue has been brought with LSTM where the recurrent cell architecture is containing a forget gate triggering a forget mechanism for each input time series, and that is learnt from the training data. The memory is therefore dynamic, and adaptive for each input time series. The attention mechanism is bringing the improvement of LSTM to a higher level as it is learning to focus, for each input time series in a temporal window, on past time ranges relevant to make a prediction [11, 12, 14]. Attention mechanism is very popular in Natural Language Processing (NLP) for translation quality improvement [15]. By analyzing the attention of a recurrent architecture, it is possible to understand where the model is focusing its attention to generate a prediction, therefore it enables to interpret temporally any predictions provided by a recurrent model [16, 17].

The proposed approach in this paper has been developed to answer the need to understand the predictions of the hot metal temperature from a blast furnace by a multivariate time series recurrent deep learning model. Due to the complex temporal and inter time series relations, several requirements are defined for the interpretability of the predictions:
- *Local interpretability*: each prediction must be explained
- *Spatial interpretability*: importance of each input variable for the prediction
- *Temporal interpretability*: for each input variable, time location in the recent history where values for that variable are important for the prediction. The time location for each input variable is corresponding to the temporal shift inherent to the blast furnace process characterized by a high inertia.

Those requirements have been covered by very few architectures available today in the literature. A CNN-based approach has been proposed in [16] where a Convolutional Neural Network (CNN) trained on multivariate time series data, is providing spatio/temporal interpretability by implementing grad-CAM [18] to generate saliency maps. However, this solution is not improving the predictive performance of the CNN model, on the contrary to attention mechanism for a recurrent-based model. In [19], a modified internal structure of LSTM is investigated and aiming at simultaneously improving the model performance and achieving model spatio/temporal interpretability. An attention-based approach for recurrent models has been proposed in [20] targeting the improvement and interpretability of the predictions. An attention mechanism is implemented to compute weights for the input features, as well as a second temporal-based attention on a LSTM model. The RETAIN model [14] brakes also the tradeoff between accuracy and interpretability of recurrent models by implementing a two-level neural attention model.

Our approach is based on the implementation of the attention mechanism in a LSTM architecture trained to predict the hot metal temperature produced by a blast furnace. Our contribution is the implementation of a dynamic attention mechanism for multivariate time series providing temporal interpretability of predictions as well as a spatial interpretability without introducing unnecessary additional parameters. The prediction interpretability is local as depending on the learnt context captured by the dynamic attention mechanism and characterizing the time changing operation of the blast furnace. A temporal attention for each prediction can be directly extracted from the state-of-the-art implementation of attention mechanism for time series. However, for our model to handle multivariate time series, a dedicated guided backpropagation has been implemented to derive the temporal attention for each individual input time series. This feature is motivated by the second and third requirements: to provide spatial and temporal interpretability.

In order to validate the interpretability provided by the spatio/temporal attention locally available for each prediction, an artificial dataset has been generated. This dataset is defined to reflect the main characteristics of the blast furnace data, namely a reaction time of some input time series on the target to predict, and therefore inducing temporal shifts in the dataset, but also variables having different level of correlation with the target.

In the next section, the deep learning architecture of the proposed approach is described. Results are presented on an artificial dataset, as well as on actual data from a blast furnace. Conclusion and perspectives of this research are discussed.

## 2 Description of the proposed approach and results

The temporal local attention mechanism presented in this paper is implemented by the architecture in Figure 1. The $n$ input time series are augmented by time series generated by a one-dimensional convolutional layer (*conv1d*) in order to learn relevant transformation of the $n$ time series. That architecture is then featuring a LSTM layer that is generating one hidden state $h_i$ for each time step $i$ in $[t-w,…, t]$ of the multivariate time series **X** concatenated with the output of *conv1d*, in a time window of size $w$. A dynamic temporal attention is applied to $w-1$ previous hidden states and the context vector $v_t$ is calculated as described in Figure 1. The dense layer of the attention mechanism bloc learns the context defined locally by the $n$ time series denoted by **X**. This context can be a specific operation mode of the blast furnace. Learning the context allows to generate dynamic attentions weigths **α**={ $α_{t-w}$, $α_{t-w+1}$, …, $α_{t-1}$ }, providing a local temporal interpretability for the prediction $y_{pred,t+horizon}$.

A dense layer generates a prediction $y_{pred,t+horizon}$ for the hot metal temperature at a horizon of 3h using as input the concatenation of the current hidden state $h_t$ and the context vector $v_t$.

A guided backpropagation-based approach is applied for each time step $t$ between each input hidden state $h^a_i$ modified by the attention mechanism, and the original input vector $x_i$ where $i$ is in $[t-w, …, t]$. The objective is to highlight which time series in the input vector $x_i$ is inducing a gradient in the hidden state $h^a_i$.

Attention Mechanism for Multivariate Time Series Recurrent
Model Interpretability Applied to the Ironmaking Industry

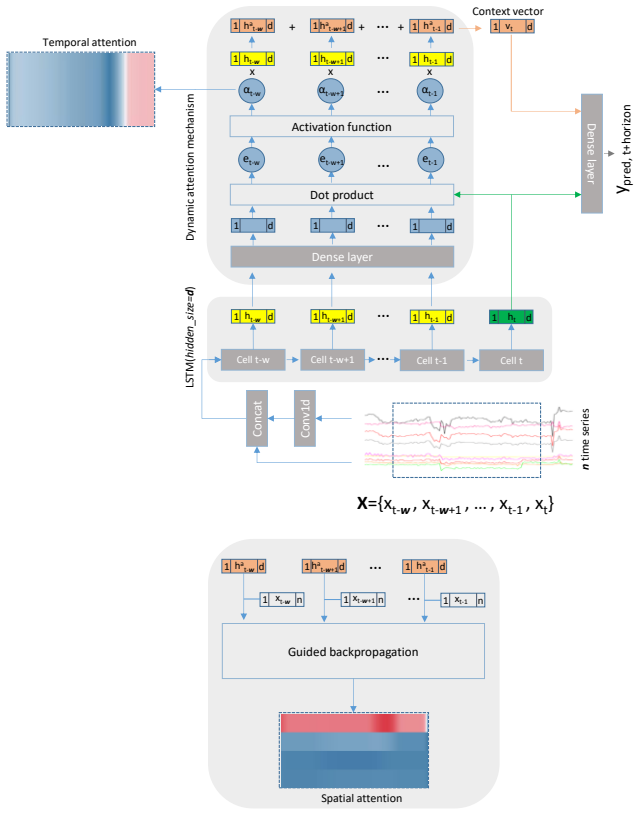

**Figure 1: Multivariate time series local spatio/temporal attention mechanism architecture**

The first validation of the proposed architecture is achieved with an artificial dataset. The objective of that validation is to ensure that the spatio/temporal attentions defined in that dataset are correctly identified. The simulated data are multivariate time series composed of four time series being inter-correlated, some with fix temporal shift, and one time series is uncorrelated to others. The following time series have been defined for creating this artificial dataset:

Time series A: $A(t)$
Time series B: $B(t) = \alpha * A(t - 3h)$
Time series C: $C(t) = \beta * B(t)$
Time series D: $D(t) = 1450$

The signal $A(t)$ is generated to randomly change its amplitude in a range of time [0, 3h]. When a change of amplitude is triggered, the amplitude itself is randomly changing in a range ±[5, 50]. The transition period of $A(t)$ is linear when a random amplitude modification occurs. Figure 2 illustrates this artificial data set. The attention-based architecture described in Figure 1 is trained on 10000 generated samples with a simulated time granularity of 1 minute, and using a time window of size $w = 500$ minutes. The parameters $\alpha$ and $\beta$ are respectively 0.1 and 0.5 with 1450 as initial value for $A(t)$. The model predicts the time series $B(t)$ from the other time series with a horizon of 3h. Figure 2a,b present the global temporal and spatial attentions calculated by averaging the local corresponding attentions over the training period. Figure 2c,d,f show the local attentions for a window of 500 minutes where a higher attention on $A(t)$ has been correctly identify on two past gradients in order to make a prediction of $B(t + 3h)$ which is as expected from the definition of the artificial dataset. The last row corresponding to the uncorrelated time series $D(t)$ has low attention without any gradient validating that no specific attention should be put on that time series. Figure 2e illustrates the evolution of $B(t)$ and its prediction.

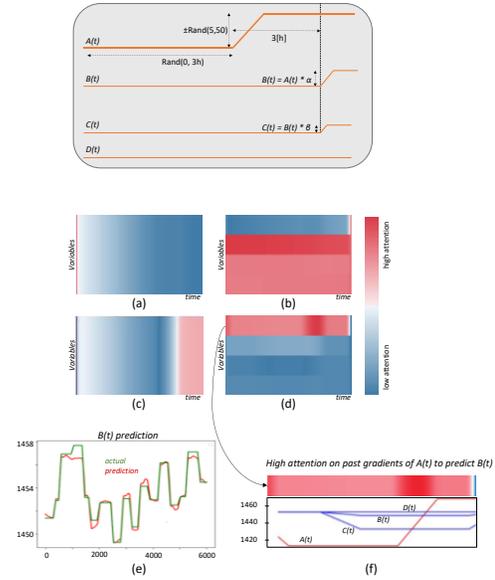

**Figure 2: Artificial dataset prediction and interpretability: (a) global temporal attention, (b) global spatio/temporal attention, (c) local temporal attention, (d) local spatio/temporal attention, (e) $B(t)$ evolution and prediction at 3h horizon**

Results on actual data are presented in Figure 4, where the model has been trained on a period of 10 months, and with n=48 input times series being continuous process variables measured at the blast furnace. The recurrent architecture is trained with a time window of size $w = 500$ minutes and that corresponds to the maximum reaction time of one variable to the hot metal temperature. The global temporal and spatial attention on the training period are illustrated in Figure 4a,b. This provides a signature of the model by understanding where it has put its attention in average during the training process, and therefore provides to the data scientist the timewise variable importance. The local temporal and spatial attention, are presented in Figure 4c,d. Those results are showing temporal and spatial attentions that are in line with known causal and correlation relations of the underlying blast furnace process.

The prediction error of the attention-based architecture is compared with the state-of-the-art implementation of a recurrent model, a Vanilla LSTM configured similarly to the attention-based architecture of this paper: same training period, same time window size $w$, same horizon of prediction. For that purpose, the Root Mean

Square Error (RMSE) of both models is calculated on one month of data, and results are summarize in Figure 4f. As expected, the attention-based architecture is improving the RMSE of a standard LSTM without attention mechanism. However, improving the RMSE is not the main objective of this research focused merely on the interpretability of the model.

interpretability), improving the attention-based architecture to reduce further the prediction error is the objective for the next phase of this research. For that purpose, specific investigation is required by using convolutional layers potentially in combination with autoregressive filters, and to design specific filters in collaboration with process experts, taking into account the physicochemical nature of the reactions happening in a blast furnace. New encoder-decoder-based architectures will be also researched to that end.

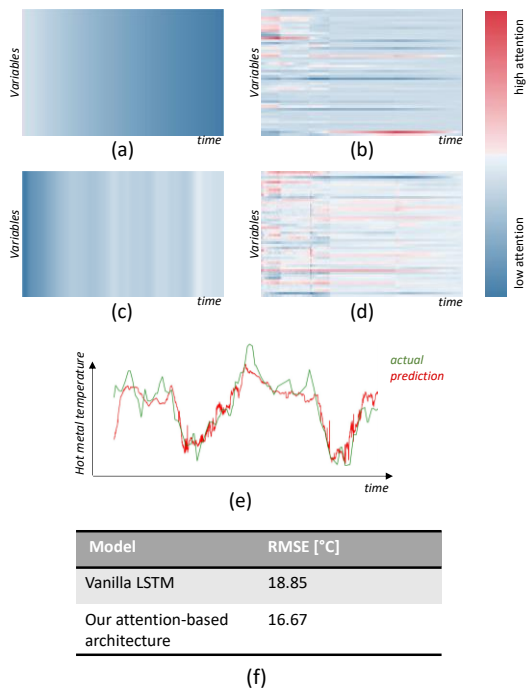

**Figure 3:** Hot metal temperature prediction and interpretability: (a) global temporal attention, (b) global spatio/temporal attention, (c) local temporal attention, (d) local spatio/temporal attention, (e) hot metal temperature evolution and prediction at 3h horizon, (f) RMSE comparison of Vanilla LSTM and attention-based model on 1 month of testing data

## 3  Conclusion and perspectives

Our results are showing a high potential in using attention mechanisms for local spatio/temporal interpretability of a multivariate time series model for the temperature prediction of the hot metal produced by a blast furnace. The attention is clearly highlighting the complex spatio/temporal relation between process variables and the hot metal temperature. Indeed, temporal shifts between actuators and hot metal temperature are resulting from the high inertia of the process in a blast furnace. This is confirmed by a dedicated validation performed on an artificial dataset that is defined to include the main features present in the blast furnace process data.

The prediction error of the attention-based model is compared with a Vanilla LSTM for the hot metal temperature prediction, by calculating the RMSE on one month of test data, which showed improvement by 12%. Although the initial motivation is to propose a new architecture with attention mechanisms, ensuring that our requirements for interpretability are met (local and spatio/temporal